\documentclass[journal]{IEEEtran}
\ifCLASSOPTIONcompsoc
  \usepackage[nocompress]{cite}
\else
  \usepackage{cite}
\fi
\usepackage{ragged2e}
\ifCLASSINFOpdf
  \usepackage[pdftex]{graphicx}
\else
  \usepackage[dvips]{graphicx}
  \DeclareGraphicsExtensions{.eps}
\fi
\usepackage{amsmath}
\usepackage{graphicx,subfigure}
\usepackage{multirow}
\usepackage{booktabs}
\usepackage{algorithm}
\usepackage{algorithmic}
\usepackage{graphicx}
\usepackage{epstopdf}
\usepackage{array}
\usepackage{amsfonts,amssymb}
\ifCLASSOPTIONcompsoc
 \usepackage[caption=false,font=footnotesize,labelfont=sf,textfont=sf]{subfig}
\else
 \usepackage[caption=false,font=footnotesize]{subfig}
\fi
\usepackage{fixltx2e}
\usepackage{dblfloatfix}
\usepackage{url}
\usepackage{bbm}
\usepackage{amsmath,amsfonts,amssymb,amsthm}
\usepackage{chngcntr}

\DeclareMathOperator*{\sign}{sign}
\makeatletter
\newcommand*\bigcdot{\mathpalette\bigcdot@{.5}}
\newcommand*\bigcdot@[2]{\mathbin{\vcenter{\hbox{\scalebox{#2}{$\m@th#1\bullet$}}}}}
\makeatother
\usepackage{color,colortbl}

\begin{document}

\title{Robustness of SAM: Segment Anything Under Corruptions and Beyond}

\author{Yu Qiao,
        Chaoning Zhang,
        Taegoo Kang,
        Donghun Kim, 
        Chenshuang Zhang, \\
        and Choong Seon Hong,~\IEEEmembership{Senior, IEEE}
\IEEEcompsocitemizethanks{
\IEEEcompsocthanksitem Yu Qiao and Chaoning Zhang are with the Department of Artificial Intelligence, School of Computing, Kyung Hee University, Yongin-si 17104, Republic of Korea (email: qiaoyu@khu.ac.kr;chaoningzhang1990@gmail.com).
\IEEEcompsocthanksitem Choong Seon Hong is with the Department of Computer Science and Engineering, School of Computing,
Kyung Hee University, Yongin-si 17104, Republic of Korea (e-mail: cshong@khu.ac.kr).
\IEEEcompsocthanksitem Corresponding author: Chaoning Zhang (e-mail: chaoningzhang1990@gmail.com)
}}

\IEEEtitleabstractindextext{%
\begin{abstract}
Segment anything model (SAM), as the name suggests, is claimed to be capable of cutting out any object and demonstrates impressive zero-shot transfer performance with the guidance of prompts. However, there is currently a lack of comprehensive evaluation regarding its robustness under various corruptions. Understanding the robustness of SAM across different corruption scenarios is crucial for its real-world deployment. Prior works show that SAM is biased towards texture (style) rather than shape, motivated by which we start by investigating its robustness against style transfer, which is synthetic corruption. Following by interpreting the effects of synthetic corruption as style changes, we proceed to conduct a comprehensive evaluation for its robustness against 15 types of common corruption. These corruptions mainly fall into categories such as digital, noise, weather, and blur, and within each corruption category, we explore 5 severity levels to simulate real-world corruption scenarios. Beyond the corruptions, we further assess the robustness of SAM against local occlusion and local adversarial patch attacks. To the best of our knowledge, our work is the first of its kind to evaluate the robustness of SAM under style change, local occlusion, and local adversarial patch attacks. Given that patch attacks visible to human eyes are easily detectable, we further assess its robustness against global adversarial attacks that are imperceptible to human eyes. Overall, this work provides a comprehensive empirical study of the robustness of SAM, evaluating its performance under various corruptions and extending the assessment to critical aspects such as local occlusion, local adversarial patch attacks, and global adversarial attacks. These evaluations yield valuable insights into the practical applicability and effectiveness of SAM in addressing real-world challenges.

\end{abstract}

}
\maketitle
\IEEEdisplaynontitleabstractindextext
\IEEEpeerreviewmaketitle

\section{Introduction}\label{sec:introduction}
\IEEEPARstart{I}{ncreasingly}, foundation models~\cite{bommasani2021opportunities} have made great strides in pushing the frontiers of modern AI. In the past few years, NLP has been revolutionalized by large language models (LLMs), which are trained on abundant text corpora collected from the web. In contrast to BERT~\cite{devlin2018bert} requiring finetuning on the downstream tasks, GPT family models~\cite{brown2020language,radford2018improving,radford2019language} demonstrate strong zero-short (or few-shot) transfer performance on unseen data distributions and new tasks. The strong capability of zero-shot transfer of such text foundation models contributes to the development of various generative AI~\cite{zhang2023complete} tasks, including text generation (ChatGPT~\cite{zhang2023ChatGPT} for instance), text-to-image~\cite{zhang2023text}, text-to-speech~\cite{zhang2023audio} and text-to-3D~\cite{li2023generative}. Despite some progress like CLIP~\cite{radford2021learning,jia2021scaling,yuan2021florence}, the progress of foundation models in computer vision~\cite{he2016deep,krizhevsky2012imagenet,qiao2023mp} lags behind. Very recently, the Meta research team released the “Segment Anything" project with the goal of building a vision foundation model for segmentation.

With masked autoencoder~\cite{zhang2022survey} mimicking BERT~\cite{devlin2018bert} to provide a unified framework for self-supervised framework in NLP and vision, the success of segment anything model (SAM)~\cite{kirillov2023segment} has also been recognized by some researchers as the GPT moment for vision by using prompts~\cite{radford2019language} (such as points and boxes). In other words, the vision community might follow NLP to go on a path of adopting the foundation model through prompt engineering. Numerous projects have combined the SAM with other models for more complex tasks beyond mask prediction, demonstrating its popularity and compatibility. Its interactive prompt design introduces greater flexibility to segmentation tasks. However, concerns arise regarding its applicability to real-world scenarios, while SAM exhibits impressive zero-shot transfer performance and high compatibility with other models. Particularly, its robustness against various corruptions (such as synthetic style-transferred corruptions and real-world common corruptions), occlusions, and perturbations (such as visible patch attacks and invisible adversarial attacks) remains unclear. 

\begin{figure}[t]
    \begin{minipage}[b]{0.5\textwidth}
    \centering
    \includegraphics[width=\textwidth]{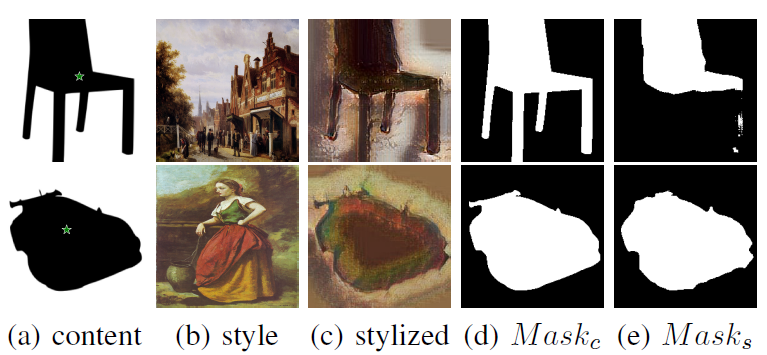}
    \end{minipage}
    \caption{The mask results of SAM in style transfer images. Figure (a) refers to the content image with the location of the point prompt marked in a green star. Figure (b) refers to the style image. Figure (c) is the stylized images after transferring style from Figure (b) to Figure (a). The white area in Figures (d), $Mask_{c}$, and (e), $Mask_{s}$, refers to the mask predicted by the given prompt, which comes from the content image and the stylized image, respectively. The results shown in Figure (e) indicate that compared to Figure (d), SAM's robustness can be affected by style images.}
    \label{fig:toy_style_transfer}
\end{figure}

In this work, we conduct a comprehensive study on the robustness of SAM in the presence of various types of corruptions and beyond. Previous research, as demonstrated by~\cite{zhang2023understanding}, has indicated that SAM tends to be biased toward texture (style) rather than shape. Inspired by this finding, our initial evaluation focuses on the robustness of SAMs to style changes. When the style is transferred to the image, it allows the image to maintain its shape so that it is still clearly visible to the human eye. Therefore, SAM theoretically should still be capable of effectively segmenting objects in such scenarios. To experimentally validate this, we provide a toy example, as depicted in Figure~\ref{fig:toy_style_transfer}. Figure (a) in Figure~\ref{fig:toy_style_transfer} represents the clean image, while Figures (b) and (c) show style images and stylized images, respectively. It is obvious that the shape of style-transferred images remains visually clear to the human eye. However, in this scenario, the predicted mask is less effective for certain stylistic variations, while it performs well for others. This contrast is particularly evident in the upper and lower subfigures in Figure~\ref{fig:toy_style_transfer} (e). This observation urges us to explore the effect of SAM on applying different style images to given content images. After a comprehensive study of different style images, we experimentally find that SAM exhibits a certain degree of robustness against style transfer. By perceiving various real-world corruptions as a new style~\cite{benz2020revisiting}, we further evaluate the robustness of SAM under various types of common corruptions~\cite{hendrycks2019benchmarking}. The common corruptions can result in a decline in image quality, often caused by factors such as noise, distortion, lighting, and other similar effects that are more prevalent in real-world scenarios~\cite{hendrycks2019benchmarking}. After investigating 15 different common corruptions (each with 5 severity levels), we observe that, apart from zoom blur, SAM maintains a certain degree of robustness against images affected by various types of noises, weather, blur, and digital distortions. This highlights that SAM is capable of executing robustly in real-world scenarios across the majority of common corruptions.

Beyond evaluating corruptions, we extend our investigation to assess SAM's robustness against local corruptions such as occlusion~\cite{ke2021occlusion} and adversarial patch attacks~\cite{brown2017adversarial}. The local occlusion refers to the intentional addition of occluding objects or elements in a specific location of an image to simulate real-world scenarios where objects might be partially obscured or blocked~\cite{naseer2021intriguing}. This type of corruption allows us to examine how well SAM can handle segmentation tasks when objects are partially hidden or covered. Our findings indicate that SAM exhibits a certain degree of robustness under various occlusion rates with 10\%, 20\%, 40\%, and 60\% of pixel values replaced with Gaussian noise. The local adversarial patch attacks involve introducing an adversarial patch that is unbounded and can arbitrarily change the pixel values at a certain location within a certain image. Moreover, it is important to consider that the employed patch is perceptible to human eyes and acts as a universal (image-agnostic) and untargeted form of attack, which can cause any input image to produce incorrect segmentation results. These properties can raise concerns that attackers can use a well-trained patch to achieve the purpose of attacking in different scenarios in reality, such as automatic driving~\cite{zhu2023understanding}. Therefore, evaluating SAM's robustness against adversarial patch attacks is crucial for understanding its ability to handle adversarial challenges that may be encountered in real-world scenarios. Specifically, we experiment with a wide range of patch sizes 0.005, 0.01, 0.02, 0.03, 0.04, and 0.05, which are defined as the proportion to the whole image, to mimic what might occur in reality. The results indicate that the adversarial patch can successfully fool the model, causing it to fail to predict the mask, even for the patch size of 0.005. Considering that adversarial patches are visible and easily detectable, we further evaluate another challenging form of image-specific adversarial attacks~\cite{goodfellow2014explaining}, which is bounded and imperceptible and typically applies to the whole image. Therefore, in this paper, we regard it as a global adversarial attack. Specifically, we adhere to common practices used in popular adversarial attack methods such as fast gradient sign method (FGSM)~\cite{goodfellow2014explaining} and projected gradient descent (PGD)~\cite{madry2017towards} attacks to assess the robustness of SAM against the global adversarial attack. The results demonstrate that SAM is vulnerable to FGSM and PGD attacks, especially for PGD attacks.

Moreover, a clear distinction is provided regarding the tasks of ``segment anything" and ``segment everything" in relation to SAM, as highlighted in~\cite{zhang2023faster}. The former involves segmenting arbitrary objects within a given image using point or box prompts, while the latter focuses on segmenting all objects present in the image. These two segmentation modes (i.e., segment anything mode which incorporates prompts such as points and boxes, and segment everything mode) offer significant flexibility across diverse image scenarios. Considering this, this work examines both segment anything and segment everything modes to evaluate SAM’s robustness under different corruption scenarios. Nonetheless, in varying scenarios of image corruptions, certain segmentation modes might be unsuitable for evaluation. For instance, in the case involving local occlusion, the ``segment everything" mode might intuitively be ill-suited for assessment, given that some pixels remain entirely obscured from view. Therefore, recognizing the distinctions between different segmentation tasks, this work adopts anything or everything mode to evaluate the robustness of SAM. To sum up, our main contributions are as follows:
\begin{itemize}
\item[$\bullet$] We are among the first to conduct a comprehensive investigation of SAM robustness across various corruptions and beyond, unveiling both its merits and limitations in handling these challenge scenarios. 
\item[$\bullet$] To the best of our knowledge, we are the first to evaluate the robustness of SAM against style change, local occlusion, and local adversarial patch attacks. 
\item[$\bullet$] Through exploring different segmentation modes, our findings indicate that SAM's robustness is inadequate against various types of attacks. Conversely, it demonstrates a certain degree of robustness against style transfer, occlusion, and common corruptions. This endeavor provides valuable insight into the efficacy of a deeper understanding of SAMs' challenges when deployed in the real world.
\end{itemize}

The remainder of this article is organized as follows. Related work on SAM is presented in Section~\ref{sec:relatedwork}. The preliminaries are introduced in Section~\ref{sec:preliminaries}. The robustness of SAM against various corruptions is evaluated in Section~\ref{sec:robust_corruptions}. Beyond corruptions, local occlusion, local adversarial patch attacks, and global adversarial attack are investigated in Section~\ref{sec:beyong_corruptions}. Finally, we conclude in Section~\ref{sec:conclusion}.

\section{Related Work}
\label{sec:relatedwork}
In this section, we first provide an overview of prior works associated with SAM in Section~\ref{subsec:sam}.
Subsequently, we review the research on adversarial attacks and their combination with SAM in Section~\ref{subsec:adv_attack}. Finally, we introduce existing works exploring style transfer, common corruptions, and occlusion, presented in Section~\ref{subsec:others}.

\vspace{-0.35cm}
\subsection{Segment anything model}
\label{subsec:sam}
In less than a month since the advent of SAM, there has been a surge in projects and papers investigating it from different perspectives. These investigations can be roughly categorized as follows. A mainstream line of research has focused on evaluating the capability of SAM to segment various objects in real-world situations accurately. Several studies have investigated its performance on different types of images, including medical images~\cite{ma2023segment,zhang2023input}, camouflaged objects~\cite{tang2023can}, and transparent objects~\cite{han2023segment}. The findings from these works consistently indicate that SAM frequently encounters difficulties in effectively detecting objects in such challenging scenarios. Another significant research direction has focused on enhancing SAM to improve its practicality. One notable approach in this line of work is Grounded SAM~\cite{GroundedSegmentAnything2023}, which incorporates text inputs to enable the detection and segmentation of various objects. This is achieved by combining Grounding DINO~\cite{liu2023grounding} with SAM, resulting in a more versatile and capable system for segmenting objects based on textual information. Another endeavor aims to extend the application scope of SAM to the mobile domain. This is achieved by substituting the heavyweight Vision Transformer (ViT)~\cite{dosovitskiy2020image} with lightweight TinyViT~\cite{wu2022tinyvit}, resulting in a significant enhancement in inference speed on mobile devices, as demonstrated in~\cite{zhang2023faster}. Due to the absence of label predictions in the generated masks by SAM, several studies~\cite{chen2023semantic,park2023segment} have sought to integrate SAM with other models such as BLIP~\cite{li2022blip} or CLIP~\cite{radford2021learning}. The objective of these efforts is to leverage the strengths of these additional models to enhance the performance and accuracy of SAM in object segmentation tasks. By combining SAM with BLIP or CLIP, researchers aim to address the limitation of SAM's generated masks without labels and achieve improved segmentation results. In addition to its applications in object segmentation, SAM has also been utilized in various other areas. Several works have explored the use of SAM for image editing purposes, including image editing techniques~\cite{rombach2022high}, as well as inpainting tasks~\cite{yu2023inpaint}. Furthermore, SAM has been employed in object tracking within videos~\cite{yang2023track,z-x-yang_2023}, demonstrating its potential for visual tracking applications. Additionally, researchers have leveraged SAM in the field of 3D object reconstruction from a single image~\cite{shen2023anything,kang2022any}, highlighting its utility in generating 3D models based on limited visual input. SAM has been reported in~\cite{tariq2023segment} to significantly improve semantic communication by only sending and receiving the foreground objects. However, most of the above works are devoted to extending the application of SAM; so far, there is a lack of comprehensive understanding of the robustness of SAM.

\begin{table}[!t]
\caption{Summary of Notations.}
\label{tab: notations}
\centering
\begin{tabular}{|c||l|}
\hline
Notation & Description\\
\hline
$SAM$ & Segment anything model\\
FGSM & Fast gradient sign method\\
PGD & Projected gradient descent\\
$\boldsymbol {x}$ & Input image\\
$y$ & Ground truth for corresponding image\\
$\widetilde {y}$ & Predicted value from corrupted image\\
$\mathcal{D}$ & dataset consisting of the image pair $\boldsymbol ({x}; y)$ \\
$\delta$ & Image perturbation \\
$\widetilde {\boldsymbol {x}}$ & $\boldsymbol {x}_{clean} + \delta$\\
$\omega$ & Model parameters\\
$\mathcal F(\cdot)$ & Classification model\\
$\boldsymbol {prompt}$ & Input prompt such as point and box\\
$\boldsymbol{mask}$ & Mask with the shape of original image\\
$mIoU$ & Averaged intersection-over-union\\
$N$ & Number of data samples\\
$precision$ & Correctly predicted pixels to total predicted pixels\\
$recall$ & Correctly predicted pixels to ground truth pixels\\
$F1$ & Combine precision and recall by harmonic mean\\
$\mathbb E(\cdot)$ & Expectation\\
$\mathcal{L}(\cdot)$ & Loss function \\
$\sign(\cdot)$ & Sign function \\
$\alpha$ & Step size for adversarial attack \\
$\rho$ & patch percentage for patch attack \\
$\epsilon$ & Upper bound for image perturbation $\delta$ \\
$\Pi(\cdot)$ & Projection function \\
$\theta$ & Threshold \\
$clip(\cdot)$ & Restrict predicted value \\
\hline
\end{tabular}
\vspace{-0.1cm}
\end{table}

\subsection{Adversarial attacks}
\label{subsec:adv_attack}
Deep neural networks (DNNs) such as CNN~\cite{qiao2023framework,goodfellow2014explaining,kurakin2016adversarial} and ViT~\cite{dosovitskiy2020image,benz2021adversarial,bhojanapalli2021understanding,mahmood2021robustness}, have demonstrated exceptional capabilities across diverse machine learning tasks, but are found to be vulnerable to adversarial examples (AEs)~\cite{szegedy2013intriguing}. The AEs are typically obtained by adding imperceptible perturbations into input images, which can deceive models, leading to inaccurate prediction outcomes. FGSM~\cite{goodfellow2014explaining} and PGD~\cite{madry2017towards} are widely used to generate AEs. This phenomenon that neural networks are sensitive to small perturbations was first systematically elaborated in~\cite{szegedy2013intriguing}, and laid the foundation for the subsequent research on adversarial attacks. Based on whether attackers can access the internal information of a model, adversarial attacks can be categorized into white-box and black-box attacks. Initial studies predominantly focus on white-box attacks~\cite{goodfellow2014explaining,madry2017towards,carlini2017towards}, where attackers have complete access to the structure and parameters of the target model. Meanwhile, researchers have also started to pay attention to more challenging black-box attacks~\cite{papernot2017practical,wang2021delving,sun2022exploring}, in which attackers can only acquire model information through limited queries. Recently, a new universal adversarial perturbation (UAP)~\cite{zhang2020cd} has been introduced, which generates a universal perturbation by traversing a large set of images. Consequently, when adding this universal perturbation to any image, it can fool neural networks to produce incorrect prediction results. Very recently, AttackSAM~\cite{zhang2023attacksam} has initially shown that SAM is vulnerable to AEs. However, these adversarial attacks are primarily concentrated in the digital domain. Given the ever-increasing deployment of deep learning systems in the real world, researchers are delving into the application of adversarial attacks in the physical domain. This technique is known as adversarial patch attacks~\cite{brown2017adversarial} and holds significant potential due to its image-agnostic nature. Different from previous studies~\cite{zhang2023attacksam,sun2022exploring,zhang2020cd,wang2021delving} that focus on one specific domain adversarial attack (either in the digital or physical domain), our work is the first to attempt to investigate the robustness of SAM from both digital and physical perspectives. We consider the image-agnostic patch attack as a local perturbation and the image-specific adversarial attack as a global perturbation. The former is visible to human eyes, while the latter is invisible to human eyes. Specifically, we first evaluate the robustness of SAM under local adversarial patch attacks. Second, We further evaluate its robustness against global adversarial attacks.

\subsection{others}
\label{subsec:others}
The seminal work in~\cite{gatys2016image} demonstrates that DNNs encode not only the content of an image but also its style information. Since then, neural style transfer (NST) has gained significant attention in both academic research and industrial applications~\cite{huang2017arbitrary, elad2017style}. Specifically, NST aims to extract stylistic information from one image using a DNN and transfer it into another content image. The drawback of the initial study~\cite{gatys2016image}, which requires solving an optimization problem, is slow. One notable method to address this issue is~\cite{johnson2016perceptual}, which uses a perceptual loss function to train a feed-forward network, resulting in real-time inference during test stages. Another significant research direction focuses on improving the universality of models so that any style images can be transferred~\cite{wang2022multi,huang2017arbitrary}. In addition, many studies~\cite{hendrycks2019benchmarking,sun2022benchmarking,yi2021benchmarking} are also devoted to evaluating the robustness of DNNs under common corruptions. A popular work has proposed a robustness benchmark to understand the vulnerability of models to corruptions~\cite{hendrycks2019benchmarking}. This has led to many works, including creating new corruptions~\cite{sun2022benchmarking,kar20223d} or applying similar corruptions for different tasks~\cite{kamann2020benchmarking,yi2021benchmarking}. Very recently, researchers have initially explored the robustness of SAM under common corruptions~\cite{huang2023robustness,wang2023empirical}. Apart from this, some methods have been proposed to restore occluded regions within the image~\cite{zhang2023bifrnet,feihong2023toward}, while others focus on evaluating model robustness against occlusion~\cite{kar20223d}. However, given that we are entering the big model moment for vision represented by SAM, a comprehensive assessment of its robustness in situations involving corruption and other situations is still lacking. Understanding its robustness in various scenarios will facilitate the comprehensive deployment of SAM in real-world applications. Table \ref{tab: notations} presents a summary of the main notations used in this manuscript.

\section{Preliminaries}
\label{sec:preliminaries}
In this section, we start by introducing the metric used to evaluate the robustness of SAM. Subsequently, we provide a preliminary overview of the implementation across different corruptions and adversarial attacks.
\subsection{Evaluation Metric}
\textbf{Mask prediction.} SAM is a model designed to perform promptable segmentation by generating masks using both images and prompts as inputs, where the prompts are the necessary component for predictions. We denote a certain input image as $\boldsymbol x$, the necessary prompts as $\boldsymbol {prompt}$, and the model parameters as $\omega$. It should be noted that the generated masks focus on object segmentation without providing semantic labels for individual masks. Therefore, for a given image, the prediction process can be formulated as follows:
\begin{equation}
    \boldsymbol{mask}_{i,j} = SAM(\boldsymbol {x}, \boldsymbol {prompt}; \omega), 
\label{eq:sam_prediction}
\end{equation}
where $\boldsymbol{mask}_{i,j}$ is the predicted mask with the shape of the original image size. The subscripts $i$ and $j$ indicate the coordinates of each pixel in the predicted mask. The pixel $\boldsymbol{x}_{i,j}$ in the original image $\boldsymbol{x}$ is considered to be within the mask area if the predicted mask $\boldsymbol{mask}_{i,j}$ for $\boldsymbol{x}_{i,j}$ is positive (greater than zero). Otherwise, it is marked as background. We denote the final predicted masks as $\boldsymbol{mask}_{predict}$, which is a binary matrix with the same shape as the original image. 

\textbf{Evaluation metric.} To quantitatively evaluate the effect of various corruptions, occlusion, and attacks on SAM, we use recall, precision, and F1-score in addition to intersection-over-union (IoU) metric commonly used in segmentation tasks. Note that for the purpose of simple description, we collectively refer to the effects of corruption, occlusion, and attacks on clean images as transformations. Specifically, we calculate the IoU between the predicted masks of a certain clean image, denoted as $\boldsymbol {mask}_{clean}$, and the predicted masks of a certain transformed image, denoted as $\boldsymbol {mask}_{predict}$. This allows us to assess the changes in masks caused by various transformations. The mIoU is then obtained by averaging the IoU scores from $N$ pairs of data samples, as shown in Equation~\ref{eq:miou}.
\begin{equation}
mIoU = \frac{1}{N}\sum_{i=1}^{N} \frac{\bigcap(\boldsymbol{mask}_{clean},  \boldsymbol{mask}_{predict})}{\bigcup(\boldsymbol{mask}_{clean},  \boldsymbol{mask}_{predict})},
\label{eq:miou}
\end{equation}
where $\boldsymbol{mask}_{(\cdot)}$ is a binary matrix indicating whether a pixel is predicted to be masked, $\boldsymbol{mask}_{clean}$ and $\boldsymbol{mask}_{predict}$ are the masked region of content images and the images after transformations, respectively. $\bigcap$ and $\bigcup$ represent the intersection and union section, respectively. Note that the mIoU score ranges from 0.0 to 1.0, with a higher value indicating more robustness.

Likewise, the remaining metrics can be defined in a similar manner. Precision is the ratio of correctly predicted pixels to total predicted pixels, which reflects the ability of the model to accurately identify the region of interest. The average precision across $N$ pairs of data samples can be defined as follows:
\begin{equation}
Precision = \frac{1}{N}\sum_{i=1}^{N} \frac{\bigcap(\boldsymbol{mask}_{clean},  \boldsymbol{mask}_{predict})}{\boldsymbol{mask}_{predict}}.
\label{eq:precision}
\end{equation}

Recall is the ratio of correctly predicted pixels to ground truth pixels, which measures the model's ability to successfully capture the actual ground truth. The average recall across $N$ pairs of data samples can be defined as follows:
\begin{equation}
Recall = \frac{1}{N}\sum_{i=1}^{N} \frac{\bigcap(\boldsymbol{mask}_{clean},  \boldsymbol{mask}_{predict})}{\boldsymbol{mask}_{clean}}.
\label{eq:recall}
\end{equation}

F1-score is a comprehensive measure that combines precision and recall through the harmonic mean, making it sensitive to cases where one of the two metrics is low. In other words, when either precision or recall is low, the F1-score will also be low. On the contrary, a high F1-score means a well-balanced prediction in terms of both precision and recall. The average F1-score across $N$ pairs of data samples can be defined as follows:
\begin{equation}
F1 = \frac{1}{N}\sum_{i=1}^{N} (2 \times \frac{Precision \times Recall}{Precision + Recall}).
\label{eq:recall}
\end{equation}

\subsection{Implementation Overview}
\label{sec:imple_details}
\textbf{Style transfer.} 
Synthetic corruption, also interpreted as style transfer, is closely related to texture synthesis and transfer~\cite{huang2017arbitrary, elad2017style}, which achieve style transfer through non-photorealistic rendering techniques~\cite{kyprianidis2012taxonomy} while preserving the overall global shape of the image~\cite{geirhos2018imagenet}. We conduct style transfer by extracting the style information from any chosen style image and applying it to a content image, resulting in the synthesis of a stylized image. The generated stylized image blends the stylistic traits of the former with the latter's content. Note that there are several algorithms available for style transfer. In this study, we employ a widely recognized and commonly used method called AdaIn~\cite{huang2017arbitrary}, which allows us to achieve arbitrary style transfer. Specifically, we perform style transfer on each content image using 5 different styles, with 45 randomly selected images for each style. This results in 225 stylized images for each content image, which enables a comprehensive evaluation of the robustness of SAM to style transfer images.

\textbf{Common corruptions.} 
To comprehensively investigate the robustness of different SAM models, in this evaluation, we utilize three distinct SAM backbones (ViT-B, ViT-L, ViT-H) to assess SAM's robustness against common corruptions. We follow the method described in~\cite{hendrycks2019benchmarking} to generate the dataset for common corruption robustness. To be more specific, we apply a range of noise corruptions, including Gaussian, shot, and impulse noise, as well as blur corruptions including glass, defocus, motion, and zoom. In addition, we consider various weather conditions such as snow, frost, fog, and brightness, along with digital corruptions such as contrast, pixelation, elastic, and JPEG compression. In total, there are 15 distinct common corruptions, each with 5 varying severity levels, where higher levels correspond to a more significant impact on the image. This results in a total of 75 corrupted images for each clean image, offering a comprehensive simulation of the expected diversity in corruption types and intensities observed in the real world.

\textbf{Local occlusion.} 
Similar to the evaluation in common corruptions, we also employ three different SAM backbones to evaluate the robustness of SAM under local occlusion. Given that various methods can be considered to define occlusion, here we adopt a straightforward approach. Specifically, we start with an image, denoted as $\boldsymbol{x} = \{ x_{i}\}_{i=1}^{N}$, where $N$ represents the total number of patches within the image. We then randomly select $M$ patches ($M < N$) and replace the pixel values of these patches with Gaussian noise. Subsequently, we evaluate how well SAM maintains mask boundaries when provided with an occluded image at a specific occlusion ratio, $\frac{M}{N}$. This empirical study provides valuable insights into mask quality for scenarios with different occlusion rates, allowing us to evaluate the reliability and utility of SAMs in the real world. Note that we evaluate the robustness of SAM across 5 distinct occlusion levels, denoted as occlusion rates $\frac{M}{N}$, which correspond to 10\%, 20\%, 40\%, 60\%, and 80\%, respectively.

\textbf{Local adversarial patch attacks.}
For typical adversarial patch attacks, attackers add a specially crafted image known as a patch to the original image, aiming to cause the model to misclassify the target image. However, for the masks generated by SAM, due to its lack of labels, an effective and direct attack method is to make SAM unable to accurately detect objects within a certain image, resulting in the removal of the generated mask after applying adversarial patches. Specifically, we specify a patch location and replace the original pixel values of that patch with randomly initialized noise, and the proportion of the patch to the entire image is defined as $\rho$. Subsequently, we optimize the patch to minimize the similarity between clean images and their corresponding images that have undergone the adversarial patch attack process. Finally, we apply the trained patch with the best attack performance to the test dataset and report evaluation results. Note that we set the range of $\rho$ to 0.005, 0.01, 0.02, 0.03, 0.04, and 0.05 for experiments. Additionally, it is worth mentioning that the optimized patches obtained through training are not image-specific; rather, they possess universality and can be applied to any test image.

\textbf{Global adversarial attacks.}
For global adversarial attacks in typical image classification models, a common goal is to add invisible perturbations to clean images, causing the classification model to produce wrong predictions. FGSM~\cite{goodfellow2014explaining} and PGD~\cite{madry2017towards} are two commonly used methods to attack classification models. The FGSM attack operates in a single step, and the PGD method can be regarded as an iterative extension of the FGSM attack, often denoted as I-FGSM~\cite{zhang2023attacksam}. However, we consistently refer to this method as PGD in this work. Inspired by the success of adversarial attacks on classification models, we adopt a similar way to evaluate the robustness of SAM again global adversarial attacks. Specifically, we adopt PGD and FGSM, aiming to effectively remove the predicted mask. Following prior research~\cite{benz2021adversarial} on attacking vision classification models in a white-box setting, we consider a range of epsilon values (here, epsilon values denote the magnitude of perturbations' intensity), denoted as $\delta = \{d / 255 \mid d \in \{0.1, 0.3, 0.5, 0.8, 1, 3, 5, 8\}\}$, for images normalized to the range of $[0, 1]$. For the PGD attack, if not specified, we adopt the PGD-20 attack, which indicates that the number of iterations is set to 20. 

\section{Robustness of SAM Against Corruptions}
\label{sec:robust_corruptions}
In this section, we evaluate the robustness of SAM under two different kinds of corruptions. The first involves style transfer, which can be regarded as a form of synthetic corruption. Next, we assess its robustness against common corruptions that might exist in the real world.

\subsection{Style Transfer}
\label{sec:style}
It has been found that SAM is more biased towards texture (style) rather than shape~\cite{zhang2023understanding}. Motivated by their finding, we first evaluate the model's robustness against style change. In contrast to~\cite{zhang2023understanding} that synthesizes an image with texture contrast on different regions, we change the texture of the whole image into a new one. Specifically, we leverage the style transfer algorithm AdaIn~\cite{huang2017arbitrary} to change the texture of the image while keeping the shape of the image content. As mentioned in the previous section, style transfer involves merging the content of one image with the style of another, creating a new image that retains the content image's shape while adopting the artistic style of the style image. We utilize a simple dataset provided in~\cite{geirhos2018imagenet} to serve as the content images to simplify our following evaluation. For style images, following the work in~\cite{westlake2016detecting}, we select 5 different ones for the experiment. 

\begin{figure}[t]
    \begin{minipage}[b]{0.5\textwidth}
    \centering
    \includegraphics[width=\textwidth]{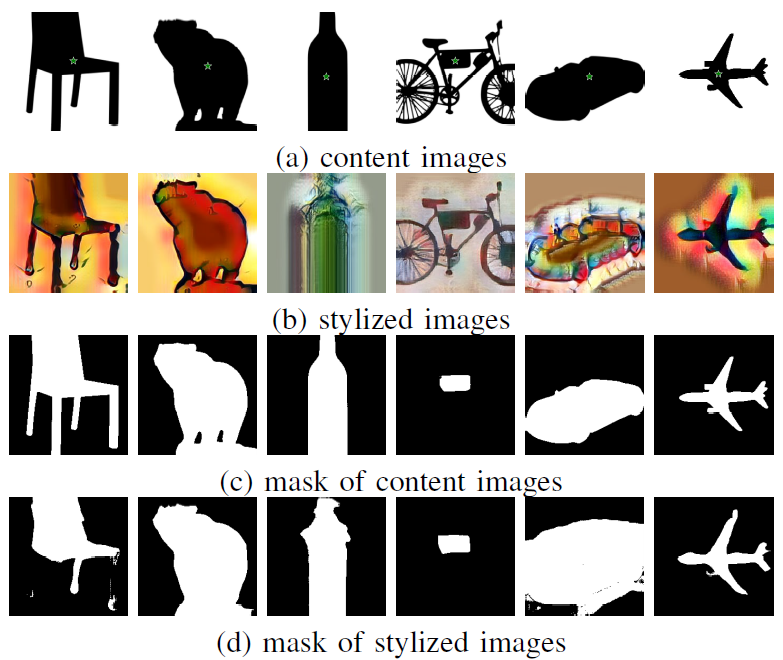}
     \end{minipage}
    \caption{Content images, stylized images, and their respective mask results. Figure (a) refers to the content image with the location of the point prompt marked in a green star and their corresponding masked results are shown in Figure (c). Figure (b) refers to the stylized images and their corresponding masked results are shown in Figure (d).}
    \label{fig:mask_ground_content}
\end{figure}

Figure~\ref{fig:mask_ground_content} displays content images and their corresponding mask results, stylized images, and their corresponding mask results. To be more specific, Figure~\ref{fig:mask_ground_content} (a) shows the content images we used for style transfer, while the mask results of them in Figure~\ref{fig:mask_ground_content} (c) can be considered as ground truth for calculating the mIoU scores after the style transfer process. Figure~\ref{fig:mask_ground_content} (b) illustrates the result of applying style transfer to different content images. It is essential to highlight that the objects within the stylized images presented in Figure~\ref{fig:mask_ground_content} (b) remain discernible and well-defined to human eyes. Given the ambiguity surrounding SAM's performance on such images, we proceed to display the corresponding mask result in Figure~\ref{fig:mask_ground_content} (d). These masks highlight areas affected by the style transfer process, providing a visual representation of changes in SAM performance. It can be observed that SAM shows a certain robustness to the style-transferred images. Moreover, we further investigate the mIoU scores across different styles, and the results are available in Table~\ref{tab:style_transfer}. The highest mIoU scores can be observed in the photorealism and suprematism styles, while the lowest scores are in the naturalism style. These findings highlight the impact of style transfer on SAM segmentation performance, revealing differences in results across styles. Nonetheless, the result in Table~\ref{tab:style_transfer} indicates that its overall level of robustness against style transfer is commendable.

\begin{table}[t]
  \caption{Style transfer mIoU results under \textbf{point} prompt.}
  \label{tab:style_transfer}
  \centering
  \resizebox{\linewidth}{!}{%
  \begin{tabular}{c|@{\hspace{4pt}}c@{\hspace{4pt}}c@{\hspace{4pt}}c@{\hspace{3pt}}c@{\hspace{3pt}}c}
    \toprule
    / & Cartoon & Naturalism & Painting & Photorealism & Suprematism\\
    \midrule
    Chair & 0.5948 & 0.4319 & 0.5872 & 0.5824 & 0.6330 \\
    Bear & 0.8867 & 0.6309 & 0.7715 & 0.8743 & 0.8766\\
    Bottle & 0.6764 & 0.6325 & 0.8040 & 0.6713 & 0.8535\\
    Bicycle & 0.7340 & 0.8051 & 0.6689 & 0.8258 & 0.6839\\
    Car & 0.6954 & 0.7497 & 0.6993 & 0.7164 & 0.7588\\
    Airplane & 0.5443 & 0.4222 & 0.5258 & 0.5591 & 0.6747\\
    \midrule
    $Average$ & 0.6886 & 0.6121 & 0.6761 & 0.7049 & 0.7468\\
  \bottomrule
\end{tabular}}
\end{table}

\begin{table}[t]
\centering
\caption{mIoU Results for Common Corruption under \textbf{everything} Mode at Severity Level 5.}
\label{tab:noise_l5}
\resizebox{\linewidth}{!}{%
\begin{tabular}{c|ccccc}
\toprule
Model & Gaussian & Shot & Impulse & Snow & Frost\\ 
\midrule
ViT-H & 0.7311 & 0.7442 & 0.7351 & 0.6950 & 0.7544 \\
ViT-L & 0.7239 & 0.7341 & 0.7256 & 0.6923 & 0.7500 \\
ViT-B & 0.6709 & 0.6856 & 0.6756 & 0.6430 & 0.7068 \\
\midrule
Model & Fog & Brightness & Defocus & Motion & Zoom\\ 
\midrule
ViT-H & 0.8334 & 0.8171 & 0.7454 & 0.7073 & 0.3739 \\
ViT-L & 0.8330 & 0.8177 & 0.7447 & 0.7077 & 0.3698 \\
ViT-B & 0.8070 & 0.7843 & 0.7166 & 0.6731 & 0.3556 \\
\midrule
Model & Glass & Contrast & Elastic & Pixelate & JPEG\\ 
\midrule
ViT-H & 0.6900 & 0.6757 & 0.7400 & 0.8835 & 0.6789 \\
ViT-L & 0.6902 & 0.6779 & 0.7426 & 0.8623 & 0.6527 \\
ViT-B & 0.6566 & 0.6351 & 0.7095 & 0.8285 & 0.5590 \\
\bottomrule
\end{tabular}}
\end{table}

\begin{table}[t]
\centering
\caption{mIoU Results for Common Corruption under \textbf{everything} Mode at all Severity Levels with ViT-H.}
\label{tab:noise_vit_h_all}
\resizebox{\linewidth}{!}{%
\begin{tabular}{c|ccccc}
\toprule
Level & Gaussian & Shot & Impulse & Snow & Frost\\ 
\midrule
1 & 0.8871 & 0.8902 & 0.8693 & 0.8334 & 0.8536\\
2 & 0.8577 & 0.8589 & 0.8386 & 0.7753 & 0.8146\\
3 & 0.8207 & 0.8268 & 0.8187 & 0.7521 & 0.7845\\
4 & 0.7824 & 0.7793 & 0.7752 & 0.6982 & 0.7841\\
5 & 0.7311 & 0.7442 & 0.7351 & 0.6950 & 0.7544\\
\midrule
Level & Fog & Glass & Defocus & Motion & Zoom\\ 
\midrule
1 & 0.8890 & 0.8741 & 0.8966 & 0.8899 & 0.5097\\
2 & 0.8710 & 0.8414 & 0.8658 & 0.8442 & 0.4562\\
3 & 0.8585 & 0.7600 & 0.8185 & 0.7905 & 0.4227\\
4 & 0.8539 & 0.7389 & 0.7806 & 0.7382 & 0.3910\\
5 & 0.8334 & 0.6900 & 0.7454 & 0.7073 & 0.3739\\
\midrule
Level & Brightness & Contrast & Elastic & Pixelate & JPEG\\ 
\midrule
1 & 0.9430 & 0.8881 & 0.6245 & 0.9469 & 0.8855\\
2 & 0.9054 & 0.8613 & 0.5152 & 0.9460 & 0.8528\\
3 & 0.8760 & 0.8244 & 0.8072 & 0.9150 & 0.8269\\
4 & 0.8450 & 0.7544 & 0.7911 & 0.9024 & 0.7588\\
5 & 0.8171 & 0.6757 & 0.7400 & 0.8835 & 0.6789\\
\bottomrule
\end{tabular}}
\end{table}

\begin{figure*}[t]
\centering
    \begin{minipage}[b]{\textwidth}
        \centering
         \includegraphics[width=\textwidth]{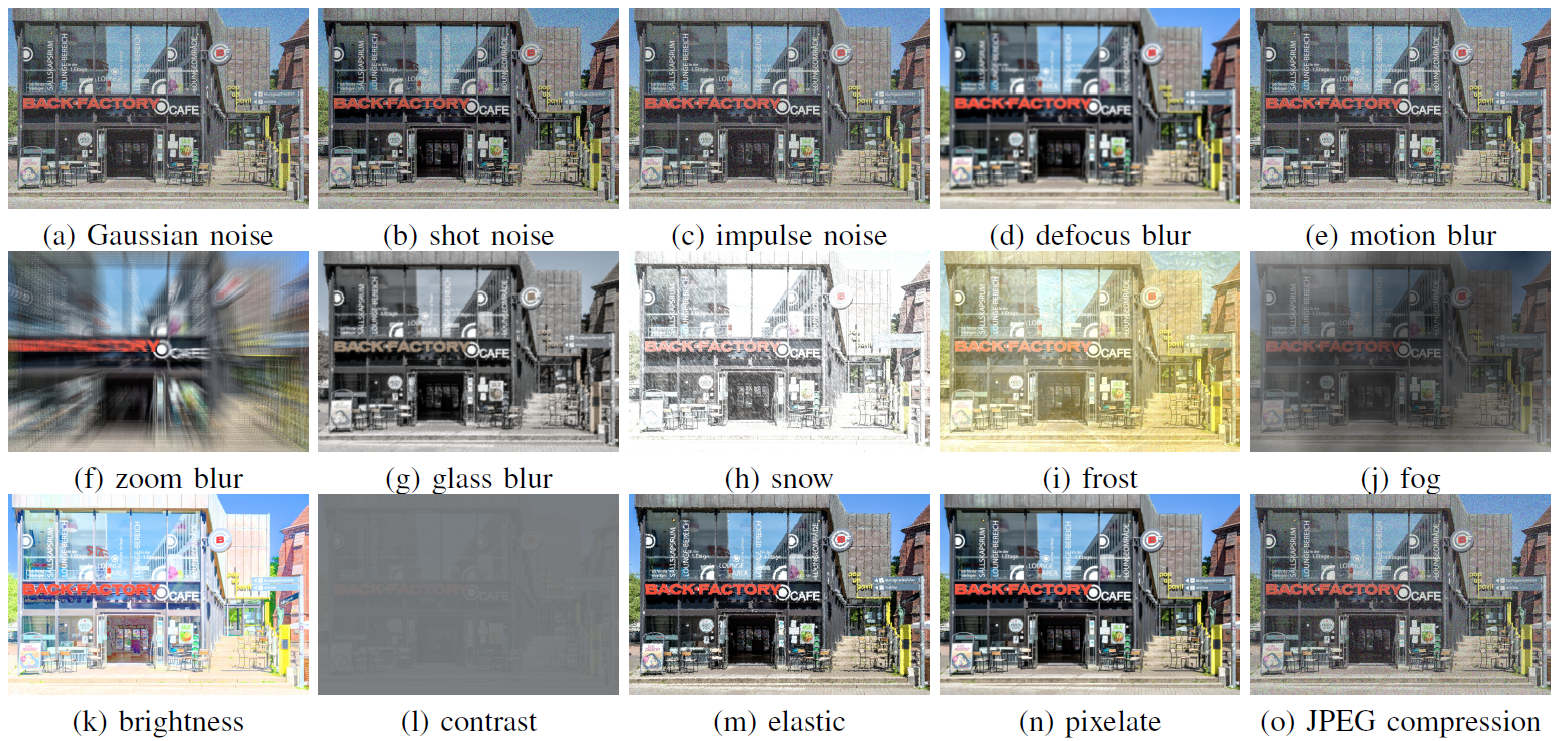}
     \end{minipage}
\caption{An example of corrupted images with severity level 5.}
\label{fig:common_corruption}
\end{figure*}

\begin{figure*}[t]
\centering
    \begin{minipage}[b]{\textwidth}
    \centering    
    \includegraphics[width=\textwidth]{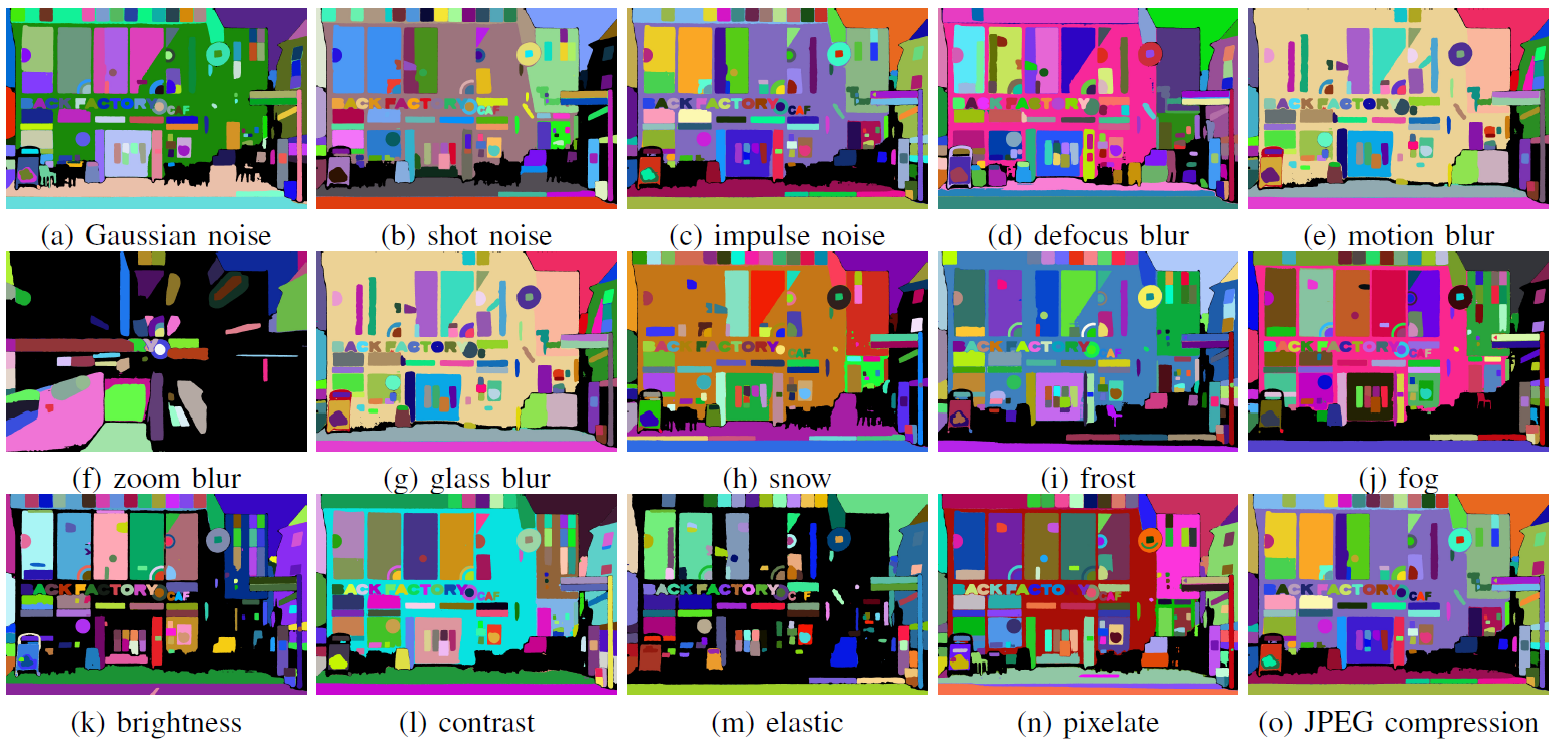}
     \end{minipage}
\caption{An example of masks of SAM for corrupted images with severity level 5.}
\label{fig:mask_common_corruptions}
\end{figure*}

\begin{figure}[t]
\centering
    \begin{minipage}[b]{0.5\textwidth}
    \centering
    \includegraphics[width=\textwidth]{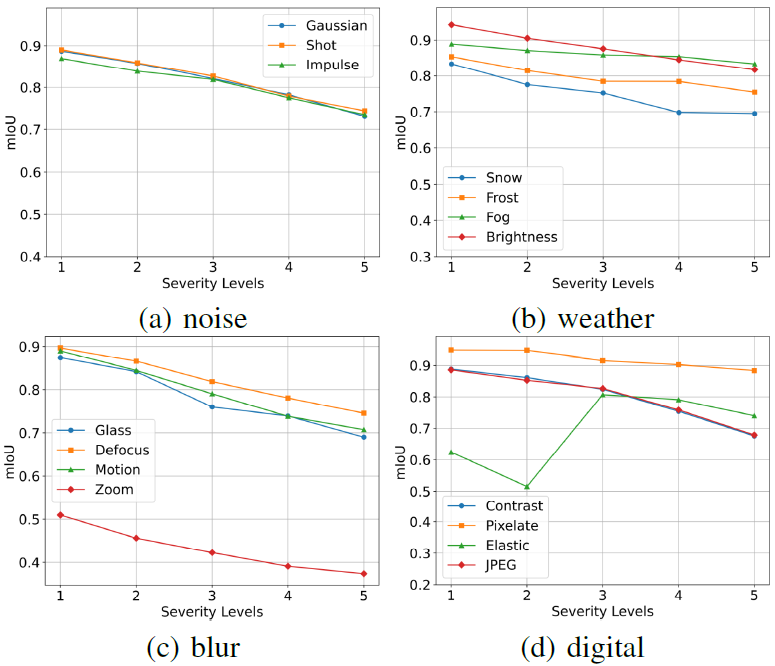}
     \end{minipage}
\caption{Illustration of the mIoU scores across various types of common corruptions. The lowest score is observed for the \textit{zoom} transformation in the blur category.} 
\label{fig:category_compar}
\end{figure}

\subsection{Common Corruptions}
\label{sec:real-world-corruption}
Various corruptions can be perceived as different new styles~\cite{benz2020revisiting}. Style change caused by the technique of style transfer in the above can be seen as a synthetic corruption that rarely occurs in practice. Therefore, we further evaluate the generalization capability of SAM under common corruptions~\cite{hendrycks2019benchmarking}. 

Specifically, to generate the dataset for assessing common corruption robustness, we start by randomly selecting 100 images from the SA-1B dataset introduced in the SAM paper~\cite{kirillov2023segment}. We evaluate the model's robustness under 15 different corruptions, each with 5 severity levels. These corruptions are broadly categorized into 4 categories: noise, digital, blur, and weather, as described in~\cite{hendrycks2019benchmarking}. In particular, the first is the noise category, which includes Gaussian, shot, and impulse noise. These types of noise typically introduce random fluctuations in pixel values due to changes within the device itself or its surroundings, resulting in the degradation of image quality. Followed by the blur category, which includes glass, defocus, motion, and zoom blur. These various types of blur can distort an image due to different factors such as focus issues, camera motion, or rapid zooming. Weather-related corruptions consist of brightness, snow, frost, and fog, which can obstruct visibility and alter the appearance of objects within the image. Lastly, the category related to digital corruption includes changes in contrast, elasticity, pixelation, and JPEG compression, which can appear with changing lighting conditions and compression techniques. Figure~\ref{fig:common_corruption} presents one example of corrupted images, demonstrating the impact of corruption at severity level 5. Figure~\ref{fig:mask_common_corruptions} provides the resulting mask generated by the ViT-H model for each corresponding image. Moreover, we adopt the segment everything mode to evaluate the robustness of SAM to common corruptions and further quantify its performance. The evaluation results are reported in Table \ref{tab:noise_l5} and Table \ref{tab:noise_vit_h_all}. Generally speaking, the results indicate that SAM exhibits high robustness against common corruptions in most cases and is likely to perform well in the real world. 

Specifically, Table~\ref{tab:noise_l5} illustrates that the mIoU score of SAM still achieves around 0.7 at the highest corruption level 5, except for zoom blur, and similar results can be observed in Table~\ref{tab:noise_vit_h_all}. The other trend is evident from Table~\ref{tab:noise_l5} that ViT-H-based SAM exhibits a higher level of robustness in comparison to ViT-B. In addition, from the result in Table~\ref{tab:noise_vit_h_all}, there is also a trend that a higher level (level 5 is higher than level 1) of corruptions corresponds to a more significant degradation in SAM performance. Intuitively, increased levels of corruption have a more significant effect on the image, therefore, observations of these trend results are in line with our expectations. Interestingly, however, as shown in Table~\ref{tab:noise_vit_h_all}, even at the lowest level (level 1), the model performs poorly for specific corruptions such as zoom blur. Moreover, we further report the mIoU scores separately according to the corruption type in Figure~\ref{fig:category_compar}. Two main observations can be drawn from the trends in Figure~\ref{fig:category_compar}, which are similar to the above findings. First, the degree of image corruption will affect the performance of SAM, but it can still maintain high robustness in most scenarios. Second, zoom blur achieves the lowest score across different corruptions, which urges researchers to design a specific model to improve its robustness. As an intuitive explanation of zoom blur, we propose that this phenomenon can be attributed to the severe distortion brought about by this corruption, which exhibits more substantial visual degradation compared to other types of corruptions, as visually depicted in Figure~\ref{fig:common_corruption} (f). Note that our results are consistent with the findings in concurrent works~\cite{huang2023robustness,wang2023empirical} that have reported the robustness of SAM in a similar setup.

\section{Beyond Corruptions}
Beyond corruptions, in this section, we first evaluate the robustness of SAM under local occlusion, and subsequently, we investigate locally visible adversarial patch attacks as well as globally invisible adversarial attacks.

\label{sec:beyong_corruptions}
\subsection{Local Occlusion}
DNN models deployed in the real world can indeed suffer from a range of the aforementioned corruptions. However, the majority of these corruptions tend to be confined to lower-level distortions (such as style changes and noise effects)~\cite{kar20223d}, which is not exhaustive for real-world evaluation. Therefore, it is critical to further explore the effect of one type of semantic distortion such as image occlusion~\cite{ke2021occlusion} on SAM performance. 

\begin{figure*}[t]
     \begin{minipage}[b]{\textwidth}
    \centering
         \includegraphics[width=\textwidth]{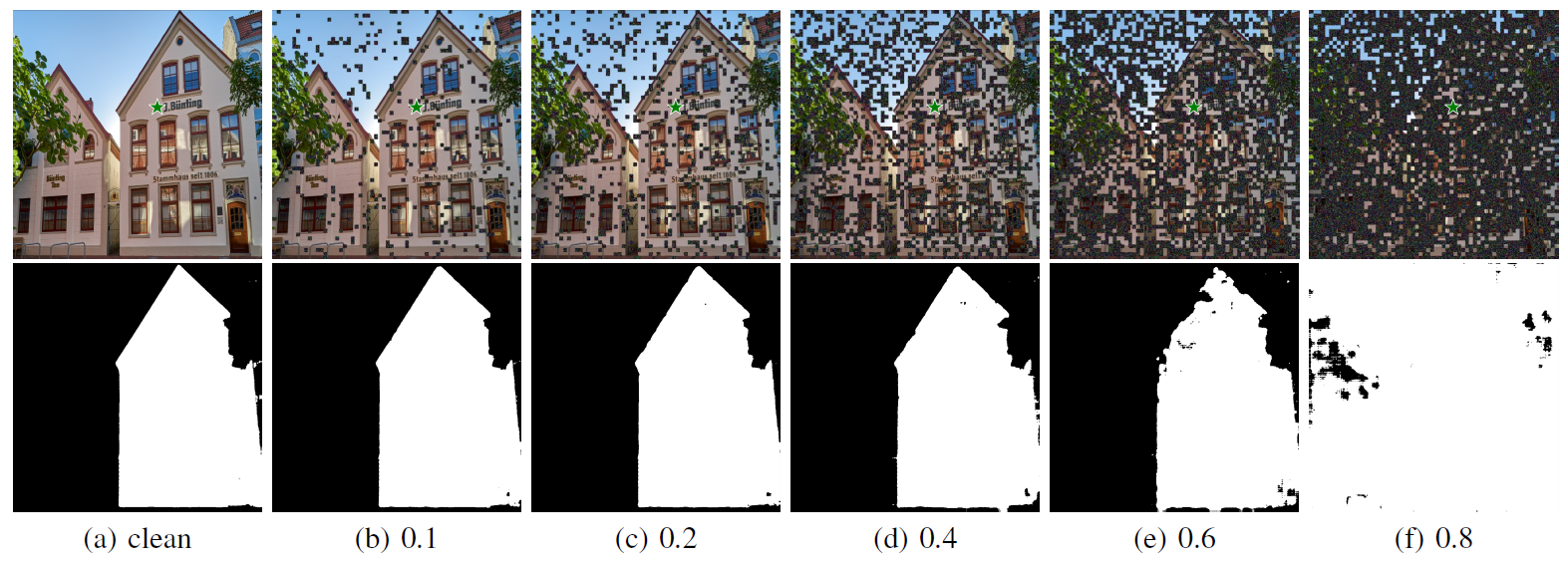}
     \end{minipage}
    \caption{An example with different levels of occlusion ratio. Figure (a) refers to the clean image without occlusion and the location of the point prompt marked in a green star. Figures (b) to (e) refer to the random occluded images with varying occlusion ratios. The pixel values of the occluded region are set to Gaussian noise with a mean of 0 and variance of 1.}
    \label{fig:mask_occlusion}
\end{figure*}

Specifically, we evaluate local occlusion using the same dataset introduced in the common corruptions section (Section ~\ref{sec:real-world-corruption}). Moreover, we apply occlusion transformations to these images, as described in Section~\ref{sec:imple_details}, thereby creating an occluded dataset tailored for comprehensive evaluation and analysis. Illustrated in Figure~\ref{fig:mask_occlusion}, an example image and its corresponding mask results. Additionally, the mIoU results, serving as a metric for local occlusion robustness evaluation, are presented in Table~\ref{tab:mIoU_occlusion}.

An overall trend can be observed from Table~\ref{tab:mIoU_occlusion} that as the occlusion ratio increases, the mIoU scores of SAM relying on various ViTs gradually decrease. This phenomenon indicates that occlusion can negatively affect the model's performance regarding segmentation accuracy. Another noteworthy observation is that SAM achieves a mIoU score of approximately 0.9 under mild occlusion (around 0.2 occlusion ratio). Moreover, even when faced with moderate occlusion (around 0.6 occlusion ratio), SAM still maintains a competitive level of performance. However, in heavily occluded scenarios with an occlusion rate of 80\%, the mIoU score experiences a significant decline. This outcome aligns with our intuitive expectation that high levels of occlusion would indeed have a significant impact on segmentation performance, given that most pixels are obscured. Furthermore, it is worth noting that ViT-H-based SAM achieves the highest mIoU values across all occlusion ratios, while ViT-B exhibits the lowest mIoU values in most cases. Overall, these findings underscore SAM's robustness across various occlusion ratios, with the exception of scenarios involving an occlusion ratio of 0.8.

\begin{figure*}[t]
    \begin{minipage}[b]{\textwidth}
         \centering
         \includegraphics[width=\textwidth]{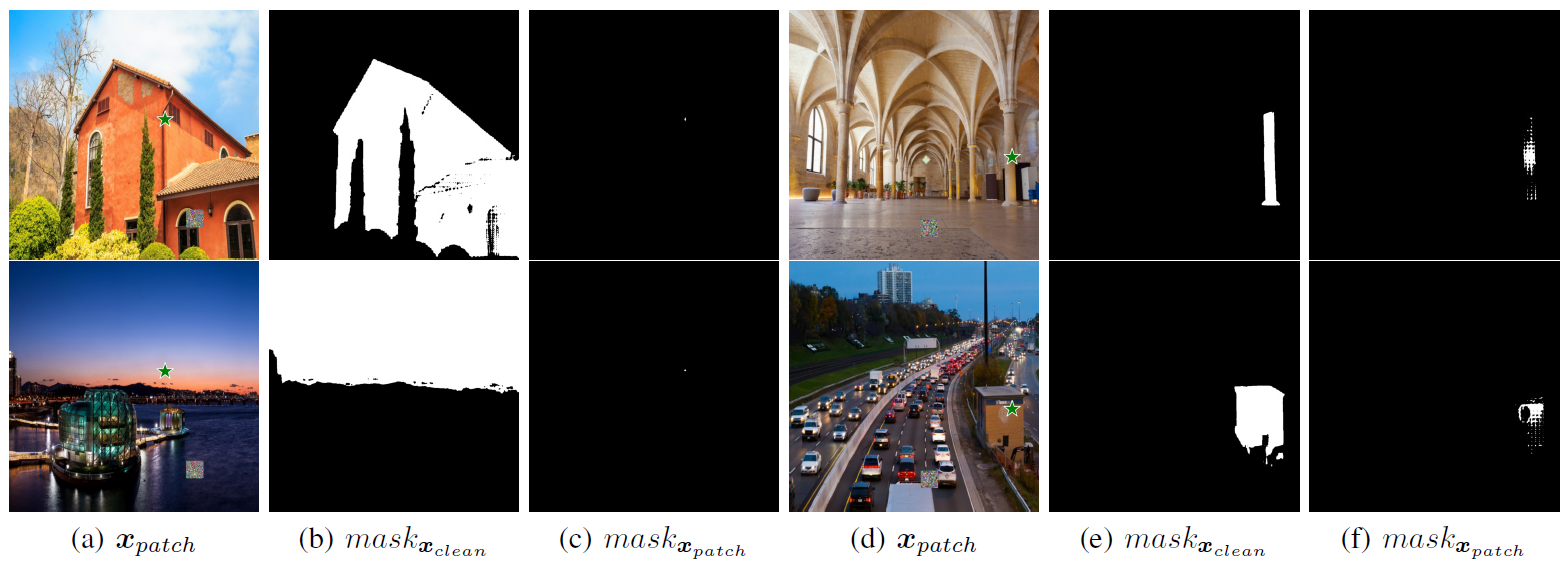}
     \end{minipage}
\caption{Examples illustrate the robustness evaluation against adversarial patch attacks with a patch percentage of 0.005 under point prompts. Figures (a) and (d) show the patched images with a green star indicating the location of the point prompt. Figures (b) and (e) display the predicted masks of clean images (without patch inside), and Figures (e) and (f) represent the predicted masks of patched images. The results in Figures (c) and (f) demonstrate that SAM is susceptible to adversarial patch attacks, as indicated by the reduced white area compared to Figures (b) and (e).}
\label{fig:mask_patch_attack}
\end{figure*}

\begin{figure*}[t] 
    \begin{minipage}[b]{\textwidth}
         \centering
         \includegraphics[width=\textwidth]{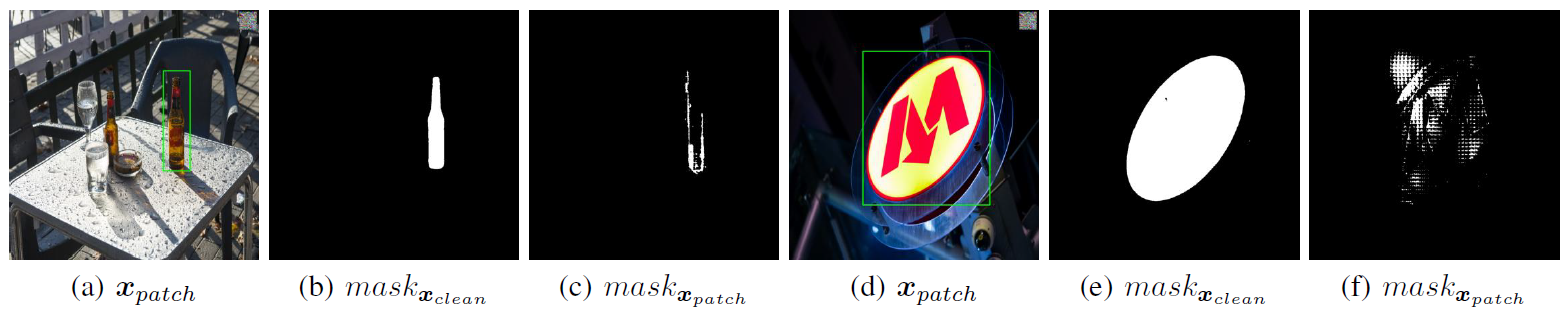}
     \end{minipage}
\caption{Examples illustrate the robustness evaluation against adversarial patch attacks with a patch percentage of 0.005 under box prompts. Figures (a) and (d) show the patched images with a green box indicating the location of the box prompt. Figures (b) and (e) display the predicted masks of clean images (without patch inside), and Figures (e) and (f) represent the predicted masks of patched images. The results in Figures (c) and (f) demonstrate that SAM is susceptible to adversarial patch attacks, as indicated by the reduced white area compared to Figures (b) and (e).}
\label{fig:box_mask_patch_attack}
\end{figure*}

\begin{table}[t]
  \caption{mIoU results for local occlusion under \textbf{point} prompt.}
  \label{tab:mIoU_occlusion}
  \centering
  \resizebox{0.90\linewidth}{!}{%
  \begin{tabular}{c|ccccc}
    \toprule
    Model & 0.1 & 0.2 & 0.4 & 0.6 & 0.8\\
    \midrule
    ViT-H & 0.8934 & 0.9234 & 0.6075 & 0.5655 & 0.2290\\
    ViT-L & 0.9151 & 0.8950 & 0.5434 & 0.6012 &0.1039 \\
    ViT-B & 0.8775 & 0.8996 & 0.5229 & 0.4696 & 0.2266\\
  \bottomrule
\end{tabular}}
\end{table}

\begin{figure*}[t]
    \begin{minipage}[b]{\textwidth}
         \centering
         \includegraphics[width=\textwidth]{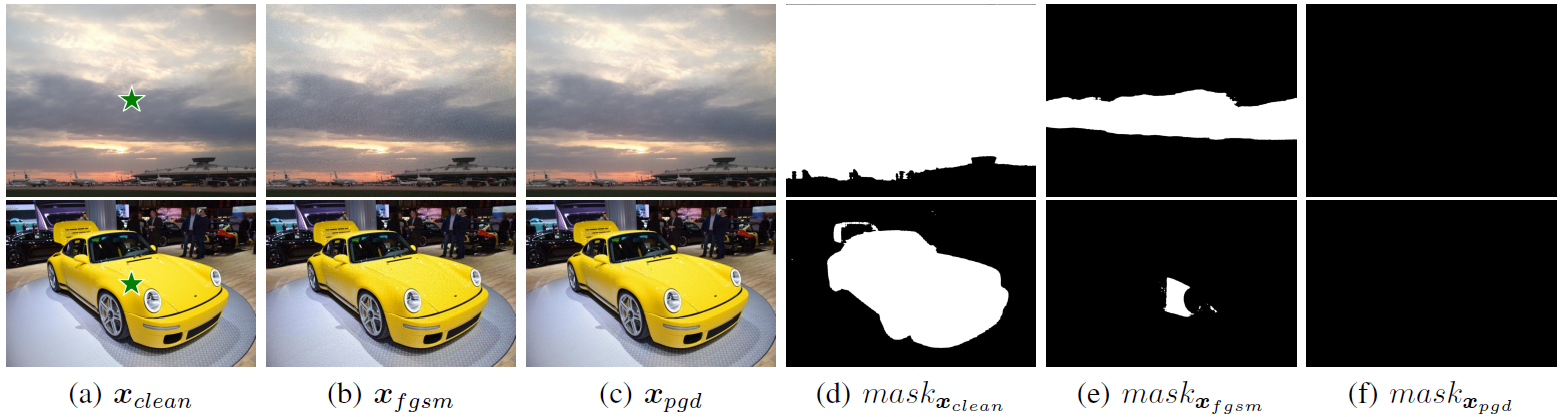}
     \end{minipage}
\caption{Examples illustrate the robustness evaluation against FGSM and PGD-20 attacks with $\delta = 8.0 / 255$ under point prompts. Figure (a) shows the clean image with a green star indicating the location of the point prompt. Figures (b) and (c) display the adversarial images generated using FGSM and PGD attacks, respectively. The white areas in Figures (d), (e), and (f) represent masks predicted by SAM based on the provided point prompt and the images from Figures (a), (b), and (c), respectively. The results in Figures (e) and (f) demonstrate that SAM is susceptible to adversarial attacks, as indicated by the reduced white area compared to Figure (d).}
\label{fig:mask_basic_attack}
\end{figure*}

\begin{figure*}[t]
    \begin{minipage}[b]{\textwidth}
         \includegraphics[width=\textwidth]{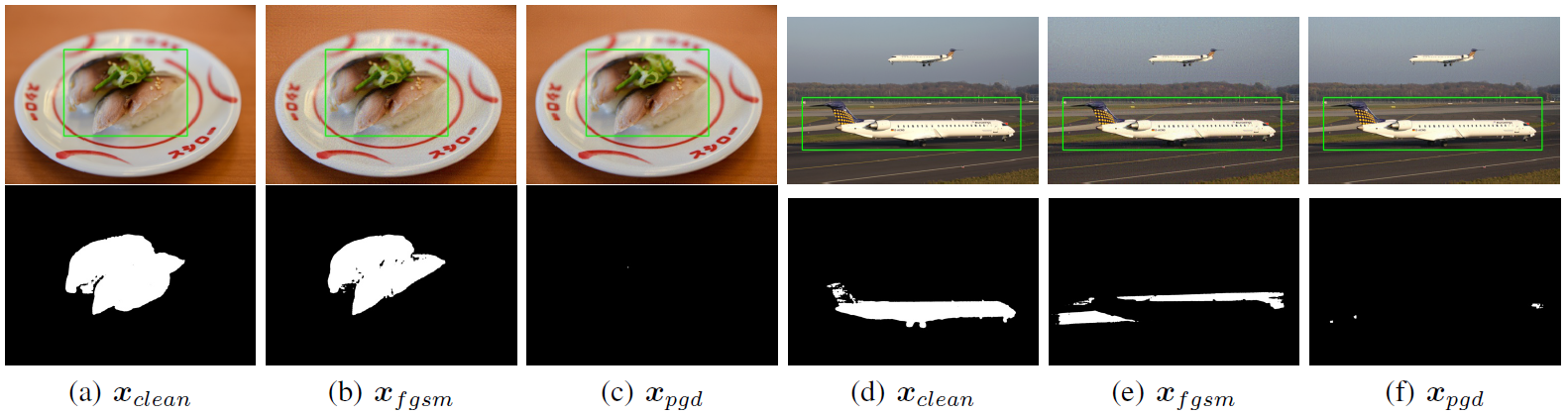}
     \end{minipage}
\caption{Examples illustrate the robustness evaluation against FGSM and PGD-20 attacks with $\delta = 8.0 / 255$ under box prompts. Figures (a) and (d) show the clean image with a green box prompt and its corresponding mask. Figures (b), (c), (e), and (f) display the adversarial images generated using FGSM and PGD attacks and their corresponding mask, respectively. The white areas from Figure (a) to (f) represent masks predicted by SAM based on the provided box prompt. The white area in Figures (b), (c), (e), and (f) demonstrate that under box prompt, SAM is still susceptible to adversarial attacks, as indicated by the reduced white area compared to Figures (a) and (d).}
\label{fig:box_mask_basic_attack}
\end{figure*}

\begin{table*}[t]
  \caption{mIoU Results for Global Adversarial Attack under \textbf{point} prompt with different ViT Models.}
  \label{tab:mIoU_attack}
  \centering
  \resizebox{0.90\linewidth}{!}{%
  \begin{tabular}{c|c|cccccccc}
    \toprule
    Model & $\delta$ & 0.1 / 255 & 0.3 / 255 & 0.5 / 255 & 0.8 / 255 & 1.0 / 255 & 3.0 / 255 & 5.0 / 255 & 8.0 / 255\\
    \midrule
    \multirow{2}{*}{ViT-H} 
& FGSM  & 0.7927 & 0.6730 & 0.5765 & 0.5276 & 0.5209 & 0.4770 & 0.4544 & 0.4600 \\
& PGD-20& 0.8020 & 0.5143 & 0.3121 & 0.1770 & 0.1721 & 0.0333 & 0.0161 & 0.0172 \\
    \midrule
    \multirow{2}{*}{ViT-L} 
& FGSM  & 0.7980 & 0.6613 & 0.5786 & 0.5632 & 0.5369 & 0.4802 & 0.4898 & 0.4906 \\
& PGD-20& 0.8258 & 0.5383 & 0.3353 & 0.1715 & 0.0895 & 0.0202 & 0.0152 & 0.0099 \\
    \midrule
    \multirow{2}{*}{ViT-B} 
& FGSM  & 0.8091 & 0.6298 & 0.5436 & 0.4870 & 0.4501 & 0.4063 & 0.4108 & 0.4132 \\
& PGD-20& 0.8159 & 0.5329 & 0.3328 & 0.2150 & 0.1479 & 0.0160 & 0.0059 & 0.0004 \\
\bottomrule
  \end{tabular}}
\end{table*}

\subsection{Local Image-Agnostic Adversarial Patch Attacks}
\label{subsec:patch_attack}
As previously discussed, SAM demonstrates a certain degree of robustness against various corruptions, including style change, common corruptions, and local occlusions. However, there remains uncertainty regarding its robustness against various adversarial attacks, which is directly related to its secure deployment in the real world. The adversarial attacks can be categorized into two types. The first is image-agnostic adversarial patch attacks~\cite{brown2017adversarial}, which alter only a small region within a clean image using a visible patch. The patch demonstrates universality, as they are crafted through training on multiple images, enabling them to have the capability to work on any given image. The other involves image-specific adversarial attacks~\cite{zhang2023attacksam}, which are crafted by adding invisible perturbations throughout an entire specific image. The purpose of both attacks is to fool the SAM's output. Since image-agnostic patch attacks only affect a small area in the image compared to image-specific attacks that affect the entire image. Here, we regard the patch attack as a local image-agnostic attack and the other attack as a global adversarial attack. In this section, we first evaluate the impact of patch attacks on SAM performance. Subsequently, global adversarial attacks are discussed in Section~\ref{subsec:global_attack}.

As pioneers in exploring SAM-based adversarial patch attacks, let us briefly review adversarial patch attacks in typical classification models. To begin with, we define $\mathcal F((\boldsymbol {x}, y); \omega)$ as the target model to be attacked where $y$ represents the corresponding labels for the input image $\boldsymbol {x}$. 
\begin{equation}
\label{eq:cls_delta}
\delta = \mathbb{E}_{(\boldsymbol {x}; y) \in \mathcal{D}}[\mathop {max} \mathcal{L}(\mathcal F(\widetilde {\boldsymbol {x}}; \omega); y))],
\end{equation}
where $\mathcal{D}$ is the dataset consisting of the image pair $\boldsymbol ({x}; y)$, $\mathcal{L}(\cdot)$ is the loss function (such as cross-entropy loss), $\widetilde {\boldsymbol {x}}$ defined as $\boldsymbol {x}_{clean} + \delta$ is the AE of $\boldsymbol {x}_{clean}$, and $\delta$ is the perturbation with no range limitation. The optimization objective is to find an adversarial patch that can cause misclassification by the classification model.

However, when dealing with segmentation tasks, the absence of direct label information makes the task notably more challenging than that of classification models. Despite this challenge, we attempt to preliminarily explore SAM-based patch attacks following the patch training strategy adopted in classification models. Therefore, given the unlabeled segmentation task based on SAM, Equation~\ref{eq:cls_delta} can be reformulated in a self-supervised manner as follows.
\begin{equation}
\label{eq:seg_delta}
\delta = \mathbb{E}_{\boldsymbol {x} \in \mathcal{D}}[\mathop {max} \mathcal{L}(\widetilde {y}, y)],
\end{equation}
where $y$ and $\widetilde {y}$ are the outputs of the clean image and patched image, respectively. Note that ${y}$ represents the ground truth, whereas $\widetilde {y}$ denotes the predicted value obtained from the patched image.

Specifically, we introduce a new loss function to measure the cosine similarity between the feature of the clean image and the corresponding patched image. This can be formulated as follows:
\begin{equation}
\mathcal{L}_{cos(\widetilde y, y)} = -\frac{<\widetilde y, y>}{||\widetilde y|| \cdot ||y||},
\label{eq:cos}            
\end{equation}
where the value of $\mathcal{L}_{cos(\cdot)}$ ranges from -1 to 1. Note that our optimization objective aims to push patched images and clean images apart by minimizing their similarity, resulting in SAM failing to accurately predict the mask. Therefore, the $\mathcal{L}_{cos(\cdot)}$ value of 1 indicates complete dissimilarity between the two, while -1 suggests complete similarity.

\begin{table}[t]
  \caption{mIoU, Recall, Precision, and F1-score Results for local adversarial patch attack under the \textbf{point} prompt with different patch percentages.}
  \label{tab:point_patch_attack}
  \centering
  \resizebox{0.95\linewidth}{!}{%
  \begin{tabular}{c|ccc}
  \toprule
    $\rho$ & 0.005 & 0.01 & 0.02 \\
    \midrule
    mIoU       & 0.0535 $\pm$ 0.0175 & 0.0251 $\pm$ 0.0143  & 0.0505 $\pm$ 0.0196 \\
    Recall     & 0.0619 $\pm$ 0.0122 & 0.0278 $\pm$ 0.0164 & 0.0561 $\pm$ 0.0235 \\
    Precision  & 0.7501 $\pm$ 0.2557 & 0.8428 $\pm$ 0.1215 & 0.9066 $\pm$ 0.0338 \\
    F1-score   & 0.0784 $\pm$ 0.0227 & 0.0401 $\pm$ 0.0218 & 0.0777 $\pm$ 0.0291 \\
    \midrule
    $\rho$ & 0.03 & 0.04 & 0.05 \\
    \midrule
    mIoU       & 0.0514 $\pm$ 0.0151 & 0.0501 $\pm$ 0.0208 & 0.0545 $\pm$ 0.0180 \\
    Recall     & 0.0556 $\pm$ 0.0172 & 0.0571 $\pm$ 0.0248 & 0.0655 $\pm$ 0.0273 \\
    Precision  & 0.9261 $\pm$ 0.0012 & 0.9009 $\pm$ 0.0214 & 0.8898 $\pm$ 0.0401 \\
    F1-score   & 0.0790 $\pm$ 0.0227 & 0.0801 $\pm$ 0.0325 & 0.0881 $\pm$ 0.0285\\
  \bottomrule
  \end{tabular}}
\end{table}

\begin{table}[t]
  \caption{mIoU, Recall, Precision, and F1-score Results for local adversarial patch attack under the \textbf{box} prompt with different patch percentages.}
  \label{tab:box_patch_attack}
  \centering
  \resizebox{0.95\linewidth}{!}{%
  \begin{tabular}{c|ccc}
  \toprule
    $\rho$ & 0.005 & 0.01 & 0.02 \\
    \midrule
    mIoU      & 0.2084 $\pm$ 0.0529 & 0.1041 $\pm$ 0.0084 & 0.0427 $\pm$ 0.0063 \\
    Recall    & 0.5252 $\pm$ 0.1957 & 0.1672 $\pm$ 0.0232 & 0.0619 $\pm$ 0.0120 \\
    Precision & 0.2799 $\pm$ 0.0149 & 0.3081 $\pm$ 0.0384 & 0.2667 $\pm$ 0.0236 \\
    F1-score  & 0.3067 $\pm$ 0.0673 & 0.1644 $\pm$ 0.0127 & 0.0703 $\pm$ 0.0127 \\
    \midrule
    $\rho$ & 0.03 & 0.04 & 0.05 \\
    \midrule
    mIoU      & 0.1537 $\pm$ 0.0133  & 0.0493 $\pm$ 0.0121  & 0.0835 $\pm$ 0.0222 \\
    Recall    & 0.2893 $\pm$ 0.0409  & 0.0797 $\pm$ 0.0296  & 0.1395 $\pm$ 0.0388 \\
    Precision & 0.3068 $\pm$ 0.0213  & 0.2729 $\pm$ 0.0323  & 0.2484 $\pm$ 0.0453 \\
    F1-score  & 0.2375 $\pm$ 0.0161  & 0.0844 $\pm$ 0.0214  & 0.1314 $\pm$ 0.0304 \\
  \bottomrule
  \end{tabular}}
\end{table}

\begin{figure*}[t]
    \begin{minipage}[b]{\textwidth}
         \centering
         \includegraphics[width=\textwidth]{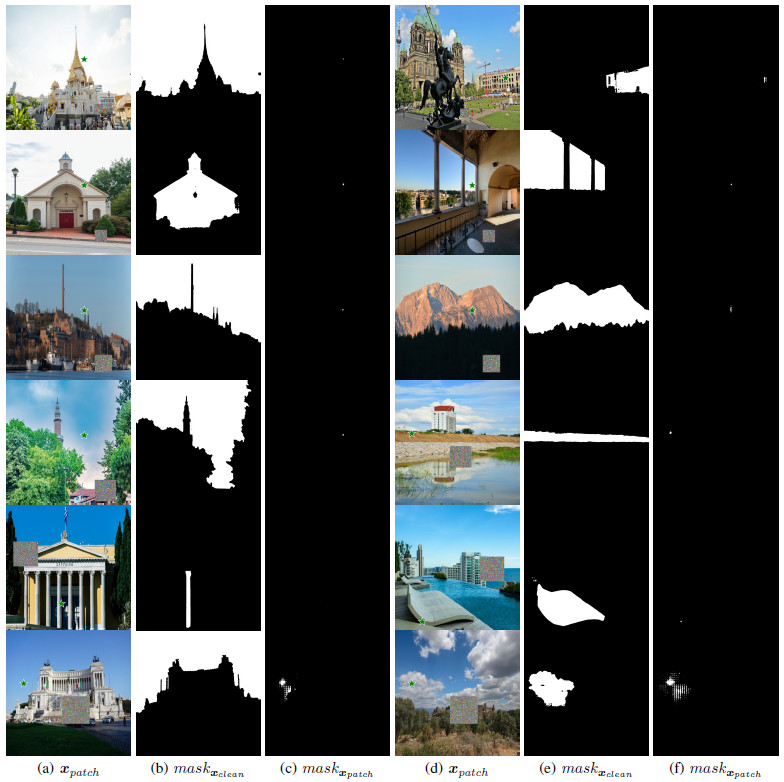}
     \end{minipage}
\caption{Examples illustrate the robustness evaluation against adversarial patch attacks with different patch percentages. The patch percentage values corresponding to the images from the first row to the last row are 0.005, 0.01, 0.02, 0.03, 0.04, and 0.05. Figure (a) shows the patched image with a green star indicating the location of the point prompt. Figures (b) and (c) display the mask result of clean images (without patch inside) and patched images, respectively. The white areas in Figures (b) and (c) represent masks predicted by SAM based on the provided prompt, respectively. In addition, Figures (d), (e), and (f) are the other corresponding image examples. The results in Figures (c) and (f) demonstrate that SAM is susceptible to adversarial patch attacks, as indicated by the reduced white area compared to Figures (b) and (e).}
\label{fig:appendx_mask_patch_attack}
\end{figure*}

For the purpose of generating a dataset to evaluate the robustness of SAM against local patch attacks, we randomly select 500 images from the SA-1B dataset introduced in the SAM paper~\cite{kirillov2023segment}. Subsequently, we evaluate the performance of the model using additional 100 test images. Without loss of generality, we conduct experiments using three different random seeds for 20 epochs each and report the mean of results and their variance. Note that considering that the mIoU score can directly reflect the attack effect, the results we report are based on the minimum mIoU score generated by each random seed and calculate the average of these minimum mIoU scores, as well as the calculation of other metrics. The optimizer for training adversarial patches is Adam, with a learning rate of 0.1. Figure~\ref{fig:mask_patch_attack} illustrates examples of patched images and their corresponding predicted masks with a patch percentage of 0.005 under point prompts. The results in Figures~\ref{fig:mask_patch_attack} (c) and (f) show that SAM is vulnerable to adversarial patch attacks, as shown by the reduction of white areas compared to (b) and (e) in Figure~\ref{fig:mask_patch_attack}. Note that additional visualization results at different patch percentages can be found in Figure~\ref{fig:appendx_mask_patch_attack}. Table~\ref{tab:point_patch_attack} shows the mean and variance of the mIoU, recall, precision, and F1-score metrics with different patch percentages ($\rho$) under the point prompt. Each patch percentage is experimented with three different random seeds. From the result in Table~\ref{tab:point_patch_attack}, there are several observations that can be made. First, patch attacks based on point prompts can fool model predictions, which can be observed by near-zero mIoU and F1 scores at different noise ratios. Second, the low variations across different locations and nearly consistent metric scores at different patch percentages indicate that patch size and location have little impact on the attack success rate. This observation can be further highlighted by the results from the Table that patches with patch percentages of 0.005 and 0.05 produce similar attack results despite a 10x size difference. Third, there is another intriguing trend of high precision but low recall, suggesting that the model's predictions can correctly include some ground-truth pixels, but perform poorly at identifying the entire object region. The illustrated results in Figure~\ref{fig:mask_patch_attack} can be further mapped to this observation.

Moreover, considering that SAM provides two kinds of prompts including points and boxes which can be used in reality, we further explore its robustness against adversarial patch attacks with box-based prompts. Figure~\ref{fig:box_mask_basic_attack} displays both the adversarial image and its corresponding mask, as well as the mask of the clean image. The results in Figures~\ref{fig:box_mask_basic_attack} (c) and (f) show that SAM still remains susceptible to adversarial patch attacks even under box prompts, as shown by the reduced white areas compared to (b) and (e) in Figure~\ref{fig:box_mask_basic_attack}. In addition, we further present the quantitative results with four metrics including mIoU, precision, recall, and F1-score in Table~\ref{tab:box_patch_attack} under different patch percentages ($\rho$), which can provide a comprehensive evaluation. Each patch percentage is experimented with three different random seeds. From the result in Table~\ref{tab:box_patch_attack}, we can observe that, similar to point-based prompt patch attacks, SAM is still vulnerable to box-based prompt patch attacks, but it is somewhat insensitive to the location and size of the patch. For example, except when the patch percentage is 0.005, the mIoU and the variances of the four metrics under different patch percentages are all around 0. In contrast to attacks based on point prompts, the precision metric under the attack based on box prompts is significantly low, as particularly evident in Figure~\ref{fig:box_mask_patch_attack} (f). The reasons behind this observation are beyond the scope of this work, however, a clear conclusion is that SAM exhibits poor robustness against both point-based and box-based prompt patch attacks.

\begin{table*}[t]
  \caption{mIoU, precision, recall, and F1-score results for global adversarial attack evaluation under \textbf{box} prompt with different ViT Models.}
  \label{tab:box_attack_evaluation}
  \centering
  \resizebox{\linewidth}{!}{%
  \begin{tabular}{c|c|cccccccc|cccccccc}
    \toprule
    \multirow{2}{*}{Model} & Metrics & \multicolumn{8}{c|}{mIoU} & \multicolumn{8}{c}{Precision} \\
    \cmidrule{2-18}
    & 255 $\times$ $\delta$ & 0.1 & 0.3 & 0.5 & 0.8 & 1.0 & 3.0 & 5.0 & 8.0 
               & 0.1 & 0.3 & 0.5 & 0.8 & 1.0 & 3.0 & 5.0 & 8.0 \\
    \midrule
    \multirow{2}{*}{ViT-H} 
& FGSM   & 0.6336 & 0.4393 & 0.3640 & 0.3171 & 0.2984 & 0.2788 & 0.2819 & 0.2764
         & 0.8541 & 0.7198 & 0.6531 & 0.6093 & 0.5943 & 0.5683 & 0.5647 & 0.5510 \\
& PGD-20 & 0.5976 & 0.2606 & 0.1233 & 0.0567 & 0.0364 & 0.0049 & 0.0059 & 0.0110
         & 0.9006 & 0.7200 & 0.5653 & 0.4479 & 0.3357 & 0.1876 & 0.2314 & 0.1488 \\
    \midrule
    \multirow{2}{*}{ViT-L} 
& FGSM   & 0.6597 & 0.4171 & 0.3224 & 0.2693 & 0.2699 & 0.2388 & 0.2465 & 0.2657
         & 0.8791 & 0.6647 & 0.5651 & 0.5057 & 0.5132 & 0.4699 & 0.4627 & 0.4830\\
& PGD-20 & 0.5703 & 0.2136 & 0.1082 & 0.0619 & 0.0366 & 0.0031 & 0.0099 & 0.0061
         & 0.8539 & 0.6318 & 0.4905 & 0.4304 & 0.3776 & 0.1949 & 0.1868 & 0.1421\\
    \midrule
    \multirow{2}{*}{ViT-B} 
& FGSM   & 0.7124 & 0.5394 & 0.4682 & 0.4249 & 0.4026 & 0.3974 & 0.4079 & 0.4003
         & 0.8688 & 0.7579 & 0.7203 & 0.6900 & 0.6677 & 0.6496 & 0.6585 & 0.6423 \\
& PGD-20 & 0.6632 & 0.3239 & 0.1625 & 0.0654 & 0.0315 & 0.0027 & 0.0012 & 0.0024 
         & 0.9043 & 0.7757 & 0.6544 & 0.4929 & 0.4139 & 0.2001 & 0.1407 & 0.1259 \\
    \midrule
    \multirow{2}{*}{Model} & Metrics & \multicolumn{8}{c|}{Recall} & \multicolumn{8}{c}{F1-score} \\
    \cmidrule{2-18}
    & 255 $\times$ $\delta$ & 0.1 & 0.3 & 0.5 & 0.8 & 1.0 & 3.0 & 5.0 & 8.0 
               & 0.1 & 0.3 & 0.5 & 0.8 & 1.0 & 3.0 & 5.0 & 8.0 \\ 
    \midrule
    \multirow{2}{*}{ViT-H}
& FGSM   & 0.6683 & 0.4957 & 0.4269 & 0.3774 & 0.3587 & 0.3401 & 0.3490 & 0.3507
         & 0.7236 & 0.5379 & 0.4613 & 0.4106 & 0.3898 & 0.3689 & 0.3745 & 0.3685 \\
& PGD-20 & 0.6119 & 0.2689 & 0.1262 & 0.0578 & 0.0372 & 0.0049 & 0.0060 & 0.0115
         & 0.6889 & 0.3382 & 0.1716 & 0.0832 & 0.0533 & 0.0092 & 0.0113 & 0.0158 \\
    \midrule
    \multirow{2}{*}{ViT-L} 
& FGSM   & 0.6874 & 0.4676 & 0.3646 & 0.3152 & 0.3200 & 0.2893 & 0.3034 & 0.3224
         & 0.7388 & 0.5119 & 0.4111 & 0.3561 & 0.3561 & 0.3179 & 0.3234 & 0.3443 \\
& PGD-20 & 0.5832 & 0.2207 & 0.1166 & 0.0624 & 0.0395 & 0.0031 & 0.0100 & 0.0062
         & 0.6507 & 0.2794 & 0.1522 & 0.0891 & 0.0572 & 0.0059 & 0.0150 & 0.0111 \\
    \midrule
    \multirow{2}{*}{ViT-B} 
& FGSM   & 0.7507 & 0.6203 & 0.5532 & 0.5171 & 0.4954 & 0.4959 & 0.5040 & 0.5031
         & 0.7936 & 0.6460 & 0.5751 & 0.5346 & 0.5093 & 0.5055 & 0.5175 & 0.5107 \\
& PGD-20 & 0.6810 & 0.3310 & 0.1655 & 0.0663 & 0.0318 & 0.0027 & 0.0012 & 0.0035
         & 0.7481 & 0.4232 & 0.2362 & 0.1033 & 0.0556 & 0.0050 & 0.0024 & 0.0042 \\
\bottomrule
  \end{tabular}}
\end{table*}

\subsection{Global Image-Specific Adversarial Attacks}
\label{subsec:global_attack}
After investigating the locally visible adversarial patch attack, it is natural to ask whether global invisible adversarial attacks can be added to the SAM. As a comprehensive investigation, we use two common attack methods, FGSM and PGD-20, to conduct experiments with different epsilon values. Note that, in both attack methods, the perturbation must be limited to a predefined range during each iteration, ensuring it is invisible, unlike patch attacks. Therefore, given the case where the perturbation is invisible, the AEs in Equation ~\ref{eq:seg_delta} can be generated by PGD as follows:
\begin{equation}
\label{eq:pgd}
\boldsymbol{x}^{t+1}=\Pi_{\boldsymbol{x} + \delta}\left(\boldsymbol{x}^t+\alpha\sign(\nabla_{\boldsymbol x}  \mathcal{L}(\widetilde {y}, y) \right),
\end{equation}
where $\alpha$ represents the step size, $\boldsymbol{x}^t$ denotes the AE generated at step $t$, $\Pi_{\boldsymbol{x} + \delta}$ represents the projection function that projects the AE onto the $\epsilon$-ball centered at $\boldsymbol{x}^0$, and $\sign(\cdot)$ denotes the sign function. Additionally, to ensure that the perturbation $\delta$ is imperceptible (or quasi-imperceptible) to the human eye, it is commonly constrained by an upper bound $\epsilon$ on the $\ell_\infty$-norm, i.e., $||\delta||_\infty \leq \epsilon$.

Given the lack of semantic labels in the SAM, we adopt a similar way as adversarial patch attacks in Section~\ref{subsec:patch_attack}, where the optimization goal is to minimize the similarity between clean and perturbed images. We follow this strategy, but for the global adversarial attack, the perturbation needs to satisfy the constraints in Equation~\ref{eq:pgd}. In addition, the global perturbation is generated by directly attacking the mask generated by SAM. Specifically, if the predicted value $\boldsymbol mask_{i, j}$ for pixel $\boldsymbol x_{i, j}$ is positive, it signifies that the pixel is masked. Therefore, the objective of successfully removing the mask is achieved when all the predicted values $mask_{i, j}$ within the designated region become negative. Here, the specific loss function in Equation \ref{eq:seg_delta} can be formulated as follows~\cite{zhang2023attacksam}:
\begin{equation}
\mathcal{L}_{mse} = -||clip(\boldsymbol{mask}, \mathrm{min}=\theta), \theta||^2,
\label{eq:cos}            
\end{equation}
where $\theta$ is a negative threshold, $clip(\cdot)$ is the operation to restrict the minimum predicted values since it makes no sense to increase the prediction values that are already below $\theta$ to be close to $\theta$. Note that we set the $\theta$ to be -10, which is also used in their work~\cite{zhang2023attacksam}. 

\begin{figure*}[t]
    \begin{minipage}[b]{\textwidth}
         \centering
         \includegraphics[width=\textwidth]{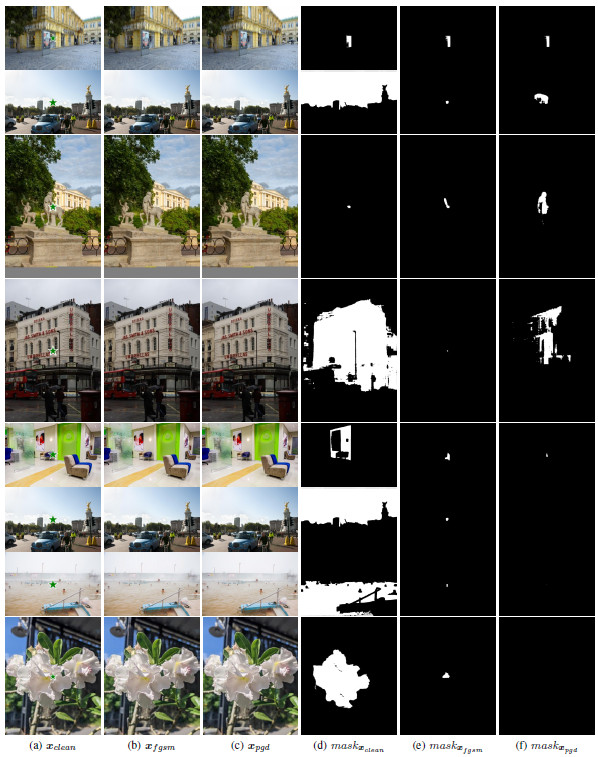}
     \end{minipage}
\caption{Robustness evaluation against FGSM and PGD-20 attacks with different values of $\delta$. The $\delta$ values corresponding to the images from the first row to the last row are as follows: 0.1, 0.3, 0.5, 0.8, 1.0, 3.0, 5.0, and 8.0. The images from left to right are as follows: clean images, images after applying FGSM, images after applying PGD, masks of clean images, masks of images after applying FGSM, and masks of images after applying PGD.}
\label{fig:appendx_mask_basic_attack}
\end{figure*}

Beyond the fixed perturbation in~\cite{zhang2023attacksam}, we perform a comprehensive evaluation on the robustness of SAM against the global adversarial attack using $\delta$ values ranging from 0.1, 0.3, 0.5, 0.8, 1.0, 3.0, 5.0, and 8.0. Visualization of clean images, adversarial images, and predicted masks is shown in Figure~\ref{fig:mask_basic_attack}. Specifically, the clean image is given a prompt indicated by a green star in Figure~\ref{fig:mask_basic_attack} (a). The generated adversarial images through FGSM and PGD attacks, as depicted in Figure~\ref{fig:mask_basic_attack} (b) and (c), respectively. The results in Figures~\ref{fig:mask_basic_attack} (e) and (f) demonstrate that SAM is susceptible to adversarial attacks, as indicated by the reduced white area compared to Figure~\ref{fig:mask_basic_attack} (d). Table~\ref{tab:mIoU_attack} presents the mIoU score for different perturbation magnitudes ($\delta$) with different ViT models. An overall trend can be observed that as the perturbation magnitude increases, the mIoU of the SAM based on different ViTs is gradually decreasing. This phenomenon indicates that SAM is also vulnerable to global adversarial attacks, especially when the attack magnitude becomes stronger. In addition, when comparing the results of FGSM and PGD, it can also be observed that the results of PGD-20 exhibit a stronger attack than that of FGSM. This finding is consistent with that in~\cite{zhang2023attacksam}, which can be attributed to the fact that PGD-20 is a stronger attack than FGSM. Overall, the results in Figure~\ref{fig:mask_basic_attack} and Table~\ref{tab:mIoU_attack} highlight the limited robustness of SAM against FGSM and PGD attacks. Particularly, PGD-20 exhibits a more significant adverse effect on the model's robustness compared to FGSM. Note that additional results with various perturbation magnitudes can be found in Figure~\ref{fig:appendx_mask_basic_attack}.

Moreover, motivated by the concerns raised in the section on local adversarial patch attacks regarding the potential for box-based attacks in real-world scenarios, we further investigate the robustness of SAM against global adversarial attacks using the box-based prompt. Specifically, following the previous step, we utilize $\delta$ values ranging from 0.1, 0.3, 0.5, 0.8, 1.0, 3.0, 5.0, and 8.0, and evaluate SAM's performance under various metrics. The visualization of two examples, including clean images, adversarial images, and corresponding predicted masks, is shown in Figure~\ref{fig:box_mask_basic_attack}. The results reveal that similar to point-based prompts, SAM remains susceptible to adversarial attacks when employing boxes as prompts, particularly in the case of PGD attacks. The quantitative results with four metrics including mIoU, precision, recall, and F1-score, are available in Table~\ref{tab:box_attack_evaluation}. We observe that SAM underperforms on all metrics, and its performance deteriorates further as the perturbation magnitude increases. For example, in the majority of cases, the mIoU score and F1-score for PGD-20 hovers close to 0, while for FGSM it is usually less than 0.3 when the perturbation magnitude reaches 3.0 / 255 and above. This finding further reveals the vulnerability of SAM to global adversarial attacks, even through boxes as prompts. However, we also observe that for the FGSM attack in Figure~\ref{fig:box_mask_basic_attack} (b), the result is not surprisingly satisfactory compared to the PGD attack in Figure~\ref{fig:box_mask_basic_attack} (c). Nonetheless, the significantly decreased performance as shown in Table~\ref{tab:box_attack_evaluation} demonstrates a successful attack against SAM, and the quantitative results have shown that SAM is not robust to global adversarial attacks.

\section{Conclusions}
\label{sec:conclusion}
In this work, we are among the early pioneers to evaluate the robustness of the SAM, for which we provide a comprehensive evaluation. Specifically, we first explore the robustness of SAM to synthetic corruptions (style transfer) with each content image investigated with 5 different styles. We find that SAM shows a certain degree of robustness against style change. Second, we investigate common corruptions that include 15 different types of corruptions in the real world. Our findings indicate that SAM is robust to common corruptions except for zoom blur. Third, we evaluate the robustness of SAM under local occlusion with different occlusion rates. We find that SAM still exhibits a certain degree of robustness under various occlusion rates except for the occlusion ratio reaching 80\%. These findings suggest that SAM is robust to style transfer, common corruptions, and occlusions to some degree. Beyond these, we further explore the effect of local image-agnostic adversarial patch attacks and global image-specific adversarial attacks on SAM under different prompts, we find that SAM exhibits limited robustness against local patch attacks and global adversarial attacks. Moreover, an intriguing phenomenon has been observed that adversarial patch attacks prove effective against SAM, but the results empirically show that patch size and location have relatively limited effects.

\bibliographystyle{IEEEtran}
\bibliography{bib_mixed,bib_local,bib_sam}

\end{document}